\documentclass[10pt,journal,compsoc]{IEEEtran}
\usepackage{amsmath,amsfonts}
\usepackage{algorithmic}
\usepackage{algorithm}
\usepackage{array}
\usepackage[caption=false,font=normalsize,labelfont=sf,textfont=sf]{subfig}
\usepackage{textcomp}
\usepackage{stfloats}
\usepackage{url}
\usepackage{verbatim}
\usepackage{graphicx}
\graphicspath{{./Figures/}}
\usepackage{cite}

\usepackage{amssymb}
\newcommand{\bbbr}{\mathbb{R}}
\DeclareMathOperator{\diag}{diag}
\newtheorem{theorem}{Theorem}
\newtheorem{remark}{Remark}

\usepackage{tikz}
\usetikzlibrary{arrows.meta, positioning, decorations.pathreplacing}

\newcommand{\Lower}[1]{\smash{\lower 1.5ex \hbox{#1}}} 

\usepackage{textcomp}
\usepackage{hyperref}
\usepackage{lipsum}

\newcommand\copyrighttext{%
  \fontsize{6.5pt}{8pt}\selectfont \textcopyright{} 2024 IEEE. Personal use is permitted, but republication/redistribution requires IEEE permission.
See https://www.ieee.org/publications/rights/index.html for more information.\vspace{3pt} \\
This article has been accepted for publication in IEEE Transactions on Pattern Analysis and Machine Intelligence. This is the author's version which has not been fully edited and content may change prior to final publication. Citation information: DOI 10.1109/TPAMI.2024.3418214}
\newcommand\copyrightnotice{%
\begin{tikzpicture}[remember picture,overlay]
\node[anchor=south,yshift=7pt] at (current page.south) {\fbox{\parbox{\dimexpr\textwidth-\fboxsep-\fboxrule\relax}{\copyrighttext}}};
\end{tikzpicture}%
}

\begin{document}
%
\title{Guaranteed Coverage Prediction Intervals with\\ Gaussian Process Regression}
%
%
%
%

\author{Harris~Papadopoulos
\thanks{H. Papadopoulos is with the Department of 
Electrical Engineering, Computer Engineering and Informatics, Frederick University, Nicosia 1036, Cyprus.\protect\\
E-mail: h.papadopoulos@frederick.ac.cy}
\thanks{Manuscript received November 25, 2022; revised October 18, 2023 and March 5, 2024; accepted June 7, 2024.}
\thanks{Digital Object Identifier 10.1109/TPAMI.2024.3418214}}

%
%

%

\markboth{IEEE Transactions on Pattern Analysis and Machine Intelligence}%
{H. Papadopoulos: Guaranteed Coverage Prediction Intervals with Gaussian Process Regression}



\IEEEtitleabstractindextext{%
\begin{abstract}
Gaussian Process Regression (GPR) is a popular regression method, which unlike most 
Machine Learning techniques, provides estimates of uncertainty for its predictions.
These uncertainty estimates however, are based on the assumption that the model is 
well-specified, an assumption that is violated in most practical
applications, since the required knowledge is rarely available. As a result, the produced uncertainty 
estimates can become very misleading; for example the prediction intervals (PIs)
produced for the 95\% confidence level may cover much less than 95\% of the true labels.
To address this issue, this paper introduces an extension of GPR based on a 
Machine Learning framework called, Conformal Prediction (CP). This extension guarantees the production 
of PIs with the required coverage even when the model is completely misspecified. 
The proposed approach combines the 
advantages of GPR with the valid coverage guarantee of CP, while the performed experimental results 
demonstrate its superiority over existing methods.
\end{abstract}

}
\maketitle
\copyrightnotice

\IEEEraisesectionheading{\section{Introduction}\label{sec:intro}}

\IEEEPARstart{T}{he} provision of confidence information about the predictions
of Machine Learning techniques is highly desirable in many practical applications, especially in safety-critical areas 
such as learning-based robotics and control~\cite{deisen:GProbocontrol,Nguyen:robotcontrol}, autonomous 
driving~\cite{kendall:drive}, estimating safe operating regions for reinforcement learning~\cite{berkenkamp:saferl} 
and medical decision support~\cite{solei:eventpred}.
Despite their importance, there are very few Machine Learning 
techniques that provide some kind of confidence information.

One such technique, which is becoming increasingly popular among the Machine Learning 
community, is Gaussian Process Regression (GPR) \cite{rasmussen:gp}. Due to its Bayesian probabilistic 
formulation, GPR produces a predictive distribution for each test instance rather than 
a plain point prediction. In addition, it has many other desirable properties, such as 
its nonparametric nature, the use of covariance functions that can be developed according to 
the data and problem in question, and the ability of using standard Bayesian methods 
for model selection. 

There is however an important drawback of the predictive 
distributions produced by GPR. They rely on the assumption of the model being well-specified. 
As a result, in the case of model misspecification (e.g. wrong GP model hyperparameters or likelihood) 
these predictive distributions can 
become very misleading. For this reason, a number of recent works, such 
as~\cite{srin:gprbounds,chow:kernelized,capone:GPRbounds,Fiedler:GPRbounds}, have attempted to
derive more reliable uncertainty bounds for GPR by scaling the posterior standard deviation produced 
for each instance. 
These however, still cannot guarantee the required coverage in all cases of model misspecification. 

This work addresses this problem by proposing an extension of the GPR approach that 
is guaranteed to produce well-calibrated prediction intervals (PIs) without assuming anything more than 
exchangeability of the data (exchangeability is a slightly weaker assumption than i.i.d.). 
The proposed approach is based on a Machine Learning framework, called 
Conformal Prediction (CP)~\cite{vovk:alrw}, that can be combined with an underlying 
technique for producing prediction regions with a provably valid coverage
of $1 - \delta$, for any significance level $\delta$.

The main building block of CP is a real-valued function, called \emph{nonconformity measure}, 
that evaluates how strange or nonconforming a candidate instance is from a set of known instances 
according to the underlying technique. In many cases it is possible to define more than one nonconformity 
measure for a given underlying algorithm and each one defines a different Conformal Predictor. 
An important attractive property of CP is that the nonconformity 
measure definition and underlying algorithm of a particular Conformal Predictor have no 
effect on its coverage guarantee, they only affect its statistical efficiency i.e. the sizes of 
the resulting prediction regions.

The CP framework has two main versions: Transductive CP (TCP), also known as Full CP, and Inductive CP (ICP), also 
referred to as Split CP. 
The former utilizes all available data for both training the underlying method and calibrating its prediction regions, 
while the latter divides the data into distinct training and calibration sets. As a result, TCP achieves higher 
statistical efficiency at the cost of increased computational demands, while ICP maintains the computational efficiency 
of the algorithm it is based on, but sacrifices some statistical efficiency. 
Additionally, there is the Cross-Conformal Prediction (CCP) version of the framework, which combines the ideas of ICP 
and cross-validation. CCP falls somewhere in between TCP and ICP in terms of statistical and computational efficiency. 
However, it is important to note that CCP does not enjoy the theoretical validity of the other two versions. 

The application of ICP to some conventional technique is rather trivial; a recent overview of  
the process can be found in~\cite{angelopoulos:gentle}. The same goes for CCP~\cite{papa:ccp}. 
On the contrary, in the case of regression the application of TCP is only possible if some 
computational trick can be employed in order to allow efficient calculation of the CP 
prediction intervals. As a result, TCP has only been applied to Ridge Regression~\cite{nouret:tcm-rr}, 
the Lasso~\cite{lei:lasso} and $k$-Nearest Neighbours Regression~\cite{papa:nnr,papa:jairnnr}. The latter also 
introduced the concept of \emph{normalized nonconformity measures} for regression CP, which result in 
improved statistical efficiency by taking into consideration the difficulty of the 
particular instance for the underlying technique. Since their introduction in~\cite{papa:nnr}, 
normalized nonconformity measures have become the standard for regression ICPs, but so far they have 
been employed in only one TCP (the $k$-Nearest Neighbours Regression TCP).

This paper proposes an algorithm for efficiently computing the PIs of TCP 
with a GPR underlying model. The proposed method combines the desirable properties of GPR 
with the \emph{model-free} coverage guarantee of CP. Furthermore, unlike all previously proposed 
TCPs apart from $k$-Nearest Neighbous Regression, it utilizes a normalized nonconformity 
measure definition based on the predictive variance produced by GPR, leading to more accurate 
PIs. 

The remainder of this paper is structured as follows. Section~\ref{sec:back} introduces the Conformal Prediction 
framework and provides an overview of relevant work. 
Following this, Section~\ref{sec:gprcp} develops two efficient versions of the transductive GPR Conformal Predictor (GPR-CP). 
One of these versions is further enhanced in Section~\ref{sec:nm2} with the introduction of a normalized nonconformity 
measure based on the predictive variance produced by GPR. Section~\ref{sec:exp} presents an experimental 
evaluation of the proposed method on both simulated and benchmark data sets, along with comparisons to other 
approaches. Finally, Section~\ref{sec:conc} 
offers concluding remarks and outlines future directions.

\section{Background}\label{sec:back}

\subsection{Conformal Prediction}\label{sec:CP}

Given a training set\footnote{The training set is in fact a multiset as it can contain some 
observations more than once.} of $l$ observations $\{z_i\}^l_{i=1}$, where each $z_i \in \mathcal{Z}$ is a pair 
$(x_i,y_i)$ consisting of the vector $x_i \in \mathcal{X}$ of inputs or features for case $i$ and the 
associated output or label $y_i \in \bbbr$ (dependent variable). Additionally, the inputs for a 
new observation $x_{l+1}$ are provided and the objective 
is to produce a prediction region that will cover the correct output $y_{l+1}$ with a 
predefined coverage rate of $1 - \delta$. As mentioned in Section~\ref{sec:intro}, the only assumption made is that all 
$(x_i, y_i)$, $i = 1, 2, \dots$, are exchangeable.

CP considers every candidate value $\tilde y \in \bbbr$ and uses the nonconformity measure to assign a value 
$\alpha^{\tilde y}_i$ to each instance $z_i$ in the extended set
\begin{equation}
\label{eq:extset}
  Z^{\tilde y} = \{z_1, \dots, z_l, z^{\tilde y}_{l+1}\},
\end{equation}
where $z^{\tilde y}_{l+1} = (x_{l+1}, \tilde y)$. Formally, a nonconformity measure is a measurable mapping 
$A:\mathcal{Z}^{(*)} \times \mathcal{Z} \rightarrow \bbbr$ (where 
$\mathcal{Z}^{(*)}$ is the set of all multisets of observations), which assigns a numerical score 
\begin{equation}
\label{eq:nmdefinition}
  \alpha^{\tilde y}_i = A(Z^{\tilde y}_{-i}, z_i)
\end{equation} 
to each $z_i \in Z^{\tilde y}$, where $Z^{\tilde y}_{-i}$ is the set resulting by the removal of $z_i$ from $Z^{\tilde y}$. 
The value $\alpha^{\tilde y}_i$, called the \emph{nonconformity score} of $z_i$, indicates how strange or 
nonconforming the pair $(x_i, y_i)$ seems in relation to the instances in $Z^{\tilde y}_{-i}$. Note that $Z^{\tilde y}_{-i}$ 
is provided to $A$ as an unordered set. 

Typically each nonconformity measure is based on some conventional Machine Learning method called 
the \emph{underlying algorithm} of the corresponding CP. Given a training set such as $Z^{\tilde y}$ each 
such method generates a prediction rule $D^{Z^{\tilde y}}$ that maps any unlabeled observation $x_i$
to a predicted label $D^{Z^{\tilde y}}(x_i)$. As this prediction rule is based on the instances in $Z^{\tilde y}$, 
a natural measure of the nonconformity of each observation $z_i \in Z^{\tilde y}$ is the degree of disagreement 
between the predicted label
\begin{equation}
\label{eq:yhat}
   {\hat y}_i = D^{Z^{\tilde y}}(x_i)
\end{equation}
and $y_i$ for $i = 1,\dots, l$ or $\tilde y$ for $i = l+1$. The simplest way of measuring the disagreement of $y_i$ and 
${\hat y}_i$ is by the absolute value of their difference $|y_i - {\hat y}_i|$.

Note that in~(\ref{eq:yhat}) $z_i$ is included in the training set 
of the underlying algorithm. Alternatively, we can generate the prediction rule $D^{Z^{\tilde y}_{-i}}$ by using $Z^{\tilde y}_{-i}$ 
as training set, and measure nonconformity as the degree of disagreement between the leave-one-out predicted label
\begin{equation}
\label{eq:yhatloo}
   \hat{\ddot{y}}_{i} = D^{Z^{\tilde y}_{-i}}(x_i)
\end{equation}
and $y_i$ for $i = 1,\dots, l$ or $\tilde y$ for $i = l+1$. Section~\ref{sec:gprcp} will consider both these cases 
with Gaussian Process Regression as underlying algorithm.

The nonconformity scores $\{\alpha^{\tilde y}_i\}^{l+1}_{i=1}$ of all instances in $Z^{\tilde y}$ 
can then be used to calculate the \emph{p-value} of the null hypothesis that $\tilde y = y_{l+1}$ as
\begin{equation}
\label{eq:pvalue}
  p(Z^{\tilde y}) = \frac{|\{i = 1, \dots, l+1 : \alpha^{\tilde y}_i \geq \alpha^{\tilde y}_{l+1}\}|}{l+1},
\end{equation} 
also denoted as $p(\tilde y)$. 

To show why $p(\tilde y)$ provides a valid p-value for testing the null hypothesis 
that $\tilde y = y_{l+1}$, note that the exchangeability assumption implies a uniform distribution over all 
permutations of $z_1, \dots, z_{l+1}$. Consequently, when
$\tilde y = y_{l+1}$, the distribution over all permutations of $\alpha^{y_{l+1}}_1, \dots, \alpha^{y_{l+1}}_{l+1}$
is also uniform, and as a result, if all $\alpha^{y_{l+1}}_i, i = 1, \dots, l+1$ are distinct $p(y_{l+1})$ is uniformly 
distributed over the set $\{1/(l+1), 2/(l+1), \dots, 1\}$. Therefore, since any equality $\alpha^{y_{l+1}}_i = \alpha^{y_{l+1}}_j$ 
for any $i$ and $j$ can only result in increasing $p(y_{l+1})$, for any significance level $\delta \in [0,1]$,
\begin{equation}
\label{eq:validity}
  \mathbb P \Bigl( (l+1) p(y_{l+1}) \leq \delta (l+1) \Bigr) \leq \delta.
\end{equation}

Based on (\ref{eq:validity}), given a significance level $\delta$, a regression 
Conformal Predictor generates the prediction region,
\begin{equation}
\label{eq:predregion}
	\mathcal{C}^{\delta}(x_{l+1}) = \{ \tilde y : p(\tilde y) > \delta \},
\end{equation}
which has a valid finite-sample coverage. This is stated in the following theorem. 

\begin{theorem}\label{thm:coverage}
(Conformal Prediction coverage guarantee~\cite{vovk:alrw}). Suppose $(x_1, y_1), \dots, (x_{l+1}, y_{l+1})$ are 
exchangeable, then the Conformal Prediction region (\ref{eq:predregion}) satisfies
$$
  \mathbb P \left( y_{l+1} \in \mathcal{C}^{\delta}(x_{l+1}) \right) \geq 1-\delta.
$$
\end{theorem}
The proof comes directly from the definition of the CP prediction region (\ref{eq:predregion}) and 
the property (\ref{eq:validity}) of the p-value function (\ref{eq:pvalue}).

Notice that the full CP framework, described here, treats both the training and the test instances in the same 
manner for calculating the p-value $p(\tilde y)$ for each candidate label $\tilde y$. Specifically, it calculates 
the nonconformity scores $\{\alpha^{\tilde y}_i\}^{l+1}_{i=1}$ by incorporating the assumed pair 
$(x_{l+1},\tilde y)$ into the training set, thus applying distinct prediction rules for each $\tilde y$. This is critical for 
ensuring a uniform distribution over all permutations of $\alpha^{y_{l+1}}_1, \dots, \alpha^{y_{l+1}}_{l+1}$, leading 
to a valid finite-sample coverage. 
In the absence of this equal treatment, the nonconformity scores of the training instances would often be biased downwards, 
resulting in prediction regions that grossly undercover.

Of course it would be impossible to explicitly consider every 
possible output $\tilde y \in \bbbr$. Section~\ref{sec:gprcp} describes how we can compute the 
prediction region (\ref{eq:predregion}) efficiently with Gaussian Process Regression as underlying algorithm.

\subsection{Related Work}\label{sec:RW}

CP was first developed as a transductive classification approach for Support Vector Machines in~\cite{gam:lbt} 
and later greatly improved in~\cite{saunders:twcc}. Soon it started being combined with several 
popular classifiers, such as $k$-Nearest Neighbours~\cite{proedrou:tcm-pr,papa:icm-pr},
Neural Networks~\cite{papa:icpnn}, Decision Trees~\cite{joh:dtcp}, Random Forests~\cite{dev:rfcp} and 
Evolutionary Classifiers~\cite{lambrou:gacp}. In all cases the developed methods were 
shown to produce well-calibrated and useful in practice confidence measures.

In the case of regression, which is the focus of this work, a computationally efficient transductive Ridge 
Regression CP was proposed in~\cite{nouret:tcm-rr}. This 
work used the absolute values of the underlying model residuals as nonconformity measure. 
Various versions of $k$-Nearest Neighbours Regression TCP and ICP along with the 
concept of normalized nonconformity measures were proposed in~\cite{papa:nnr,papa:jairnnr}.
Normalized nonconformity measures divide the standard regression measure (the absolute values of 
the underlying model residuals) with a measure of the difficulty of each instance for the underlying model, 
thus resulting in more precise PIs.
The nonconformity measures proposed in~\cite{papa:jairnnr} used the distance of 
the $k$-nearest neighbours and the standard deviation of their labels as difficulty measures. 
The transductive version of the framework was also applied to the Lasso in~\cite{lei:lasso}, with 
only the standard non-normalized measure however.

As opposed to the transductive form of the framework, where normalized nonconformity measures 
have only been developed for $k$-Nearest Neighbours Regression, in the inductive form of the framework 
various normalized nonconformity measures have been developed for other popular regression techniques 
such as Ridge Regression~\cite{papa:icm-rr}, 
Neural Networks Regression~\cite{papa:nnetricp} and Regression Forests~\cite{joh:rcprf,bost:arfcp}.
The normalized measures used in these works assess the difficulty of each instance as: (1) the 
predictions of a separate linear model trained on the residuals of the original underlying 
model~\cite{papa:icm-rr,papa:nnetricp,joh:rcprf}; (2) the average error of the $k$-nearest 
neighbours of the instance~\cite{joh:rcprf}; (3) the variance of the predictions of individual ensemble members~\cite{bost:arfcp}.
A recent study~\cite{joha:normalized} investigates the effect of different underlying techniques, difficulty measures,
and their parameters, on the performance of normalized inductive conformal regressors with 
random forest and gradient boosting. 

CP has attracted a lot of attention in application areas where the reliable quantification 
of uncertainty is important, such as ovarian cancer detection~\cite{gam:preprot}, stroke risk assessment~\cite{papa:stroke}, 
drug discovery~\cite{eklu:drug}, bioactivity modeling~\cite{sven:bioactivity}, electricity price 
forecasting~\cite{kath:power}, software effort estimation~\cite{papa:soft} and text 
classification~\cite{malt:multilabel}. It has also been extended to additional problem settings such as 
multi-label learning~\cite{lamb:multilabel,papa:multilabel}, multi-target regression~\cite{mess:ellipsoidalcopa}, 
change detection in data streams~\cite{ho:change}, concept drift detection~\cite{eliad:driftcopa}, 
anomaly detection~\cite{laxha:anomaly} and active learning~\cite{ho:active}. An extensive list of works and 
resources on CP can be found at~\cite{manokhin:awesome}.

A related approach to CP is the \emph{jackknife prediction}, which defines PIs using 
the quantiles of leave-one-out residuals. However, the jackknife method does not provide any theoretical 
guarantees and in fact can have extremely poor coverage in some cases~\cite{barber:jackknife+}. This was recently addressed 
with the introduction of \emph{jackknife+} in~\cite{barber:jackknife+}, which like CP provides coverage guarantees under the 
assumption of exchangeability. Unlike CP though, jackknife+ provides a guaranteed coverage rate 
of $1 - 2\delta$ (instead of $1 - \delta$). Furthermore, as opposed to CP with normalized nonconformity 
measures and therefore the approach proposed in this work, jackknife+ does not provide any way of taking 
into consideration the difficulty of each instance for the underlying model, which results in more precise and informative PIs.

An alternative direction to uncertainty quantification for regression involves the provision of probability distributions 
instead of PIs through post-hoc model calibration~\cite{kule:calib,song:distcal,marx:modular}. Probability distributions offer the 
advantage of conveying richer information compared to PIs and can be readily transformed into PIs for any desired significance level. 
An extension of the CP framework for producing predictive distributions that are guaranteed to be calibrated in probability, 
named Conformal Predictive Systems (CPS), was developed in~\cite{vovk:cps1st}. This approach was extended in~\cite{vovk:cps} to Inductive 
CPS that can calibrate any model for producing fully adaptive to the test object predictive distributions.

Recalibration techniques however, including Inductive CPS, have a significant drawback compared to TCP. Achieving well-calibrated 
probability distributions necessitates setting aside a substantial portion of the available data for the calibration step, similarly 
to ICPs. This results in reduced 
statistical efficiency or even in some cases poor calibration when the data is limited.

Unlike recalibration techniques and ICPs, the approach developed in this paper utilizes the entire data set for both 
training the underlying model and calibrating the generated PIs, resulting in higher statistical efficiency. Furthermore, it 
combines the attractive properties of GPR, with the coverage guarantee of TCP and the precise PIs of a normalized 
nonconformity measure. Notably, it is the second TCP to incorporate a normalized measure.

\section{Gaussian Process Regression CP}\label{sec:gprcp}

This section begins with a brief overview of Gaussian Process Regression (GPR). A detailed 
introduction to all aspects of GPR can be found in~\cite{rasmussen:gp}.
GPR assumes that each output $y_i$ is generated as $y_i = f(x_i) + \epsilon$, 
where $\epsilon$ is additive observational noise that follows a Gaussian distribution 
$\mathcal{N}(\epsilon|0,\sigma^2_n)$ and $f$ is drawn from a Gaussian Process on the input space $\mathcal{X}$ 
with covariance function $k: \mathcal{X} \times \mathcal{X} \rightarrow \bbbr$ and mean 
$\mu: \mathcal{X} \rightarrow \bbbr$. This in effect means that the joint distribution of any 
collection of function values $\mathbf{f} = (f(x_1), \dots, f(x_l))$ associated with the 
inputs $\mathbf{X} = (x_1, \dots, x_l)$ is multivariate Gaussian
\begin{equation}
p(\mathbf{f}|\mathbf{X}) = \mathcal{N}(\mathbf{f}|\mu,K),
\end{equation}
where $K$ is an $l \times l$ covariance matrix; in particular $K_{i,j} = k(x_i,x_j)$. 
For notational simplicity the mean function will be set to zero for the remainder of the paper. 

The conditional distribution of a function value $f(x_{l+1})$ given the test input $x_{l+1}$ and 
the training observations $\{z_i\}^l_{i=1}$ is also Gaussian with mean and 
variance
\begin{subequations}\label{eq:gppv}
\begin{align}
	\hat{y}_{l+1}& = \mathbf{k}_{l+1}^{\top}(K + \sigma^2_nI)^{-1}\mathbf{y}, \label{eq:gppred} \\
	\sigma^2_{l+1}& = k(x_{l+1}, x_{l+1}) - \mathbf{k}_{l+1}^{\top}(K + \sigma^2_nI)^{-1}\mathbf{k}_{l+1}, \label{eq:gpvar}
\end{align}
\end{subequations}
where $\mathbf{k}_{l+1}$ is the vector of covariances between $x_{l+1}$ and the $l$ the training inputs, 
$\mathbf{y} = (y_1, \dots, y_l)^{\top}$ is the vector of observed outputs and $I$ is the identity matrix.
The predictive distribution for the test output $y_{l+1}$ can be obtained by adding the noise 
variance $\sigma^2_n$ to~(\ref{eq:gpvar}).

The corresponding expressions for the leave-one-out cross-validation predictive mean and variance 
(see~\cite{rasmussen:gp}, equation (5.12)) for an observation $i$ are:
\begin{subequations}\label{eq:gploopv}
\begin{align}
	\hat{\ddot{y}}_i& = y_i - \frac{[(K + \sigma^2_nI)^{-1}\mathbf{y}]_i}{[(K + \sigma^2_nI)^{-1}]_{ii}}, \label{eq:gploopred} \\
	\hat{\ddot{\sigma}}^{2}_{i}& = \frac{1}{[(K + \sigma^2_nI)^{-1}]_{ii}}, \label{eq:gploovar}
\end{align}
\end{subequations}
where $[\mathbf{v}]_i$ is the $i$th element of the vector $\mathbf{v}$ and $[M]_{ii}$ is 
the $i$th diagonal element of the matrix $M$. Although this is used in~\cite{rasmussen:gp} for model selection, it is 
particularly convenient for the construction of GPR-CP with $Z^{\tilde y}_{-i}$ as training set for measuring the 
nonconformity of each $z_i \in Z^{\tilde y}$ (see (\ref{eq:yhatloo})).

The rest of this section develops a Conformal Predictor with GPR as underlying algorithm. 
As discussed in Section~\ref{sec:CP}, every CP is based on a nonconformity measure, which assigns a 
score $\alpha^{\tilde y}_i$ to every observation $z_i \in Z^{\tilde y}$ (see (\ref{eq:extset})). This score quantifies 
the degree of disagreement between the observed label $y_i$ and the predicted value $\hat y_i$ (\ref{eq:yhat}) or 
$\hat{\ddot{y}}_i$ (\ref{eq:yhatloo}).
The obvious nonconformity measure in the context of regression is
\begin{equation}
\label{eq:nm1a}
	\alpha_i = |y_i - \hat y_i|,
\end{equation}
when the underlying algorithm is trained on the whole set $Z^{\tilde y}$ and 
\begin{equation}
\label{eq:nm1b}
	\alpha_i = |y_i - \hat{\ddot{y}}_i|,
\end{equation}
when the set $Z^{\tilde y}_{-i}$ is used as training set of underlying algorithm. 
This measure can also be enhanced by normalizing it with the difficulty of each input 
for the particular underlying technique~\cite{papa:jairnnr}. One such measure is defined in the 
next Section.

Lets first consider the nonconformity measure (\ref{eq:nm1a}). By replacing $\hat y_i$ with the GPR mean~(\ref{eq:gppred}), the 
vector of nonconformity scores $|\mathbf{y} - \hat{\mathbf{y}}| = (\alpha_1, \dots, \alpha_{l+1})^{\top}$ can 
be written in the form
\begin{align*}
    |\mathbf{y} - \hat{\mathbf{y}}|& = |\mathbf{y} - K(K + \sigma^2_nI)^{-1}\mathbf{y}| \\
                                   & = |(I - K(K + \sigma^2_nI)^{-1})\mathbf{y}| \\
                                   & = |\sigma^2_n(K + \sigma^2_nI)^{-1}\mathbf{y}|.
\end{align*}
Since $\mathbf{y} = (y_1, \dots, y_l, 0)^{\top} + (0, \dots, 0, \tilde{y})^{\top}$, the vector 
of nonconformity scores can be represented as $|\mathbf{a} + \mathbf{b}\tilde{y}|$, where
\begin{subequations}\label{eq:AandB1}
\begin{align}
    \mathbf{a}& = \sigma^2_n(K + \sigma^2_nI)^{-1}(y_1, \dots, y_l, 0)^{\top}, \\
    \mathbf{b}& = \sigma^2_n(K + \sigma^2_nI)^{-1}(0, \dots, 0, 1)^{\top}.
\end{align}
\end{subequations}
As a result, the nonconformity score $\alpha_i = \alpha_i(\tilde{y})$ of each observation 
$i = 1, \dots, l+1$ is a piecewise-linear function of $\tilde y$.
Therefore, as the value of $\tilde y$ changes, the p-value $p(\tilde y)$ (defined by (\ref{eq:pvalue})) 
can only change at the points where $\alpha_i(\tilde y) = \alpha_{l+1}(\tilde y)$ 
for some $i = 1, \dots, l$. This means that instead of having to calculate the 
p-value of every possible $\tilde y \in \bbbr$, we can calculate the p-values of the finite 
number of points where $\alpha_i(\tilde y) = \alpha_{l+1}(\tilde y)$ and of the intervals 
between them, leading to a feasible prediction algorithm~\cite{nouret:tcm-rr}.

Before going into the details of the algorithm lets first perform the same transformation 
with (\ref{eq:nm1b}) as nonconformity measure. In effect this corresponds to $\hat{\ddot{y}}_i$ being the 
leave-one-out cross-validation predictions of GPR (\ref{eq:gploopred}) and therefore the vector of 
nonconformity scores  $|\mathbf{y} - \mathbf{\hat{\ddot{y}}}| = (\alpha_1, \dots, \alpha_{l+1})^{\top}$ can 
be written in the form
\begin{align*}
    |\mathbf{y} - \mathbf{\hat{\ddot{y}}}|
										& = \left| \mathbf{y} - \mathbf{y} + (K + \sigma^2_nI)^{-1}\mathbf{y} ./ \diag((K + \sigma^2_nI)^{-1}) \right| \\
                    & = \left| (K + \sigma^2_nI)^{-1}\mathbf{y} ./ \diag((K + \sigma^2_nI)^{-1}) \right|, 
\end{align*}
where $\diag(M)$ is a vector containing the diagonal elements of matrix $M$ and $\mathbf{u} ./ \mathbf{v}$ 
is the element-by-element division of the vectors $\mathbf{u}$ and $\mathbf{v}$. As a result,
the vector of nonconformity scores can be represented as $|\mathbf{a} + \mathbf{b}\tilde{y}|$, where
\begin{subequations}\label{eq:AandB2}
\begin{align}
    \mathbf{a}& = (K + \sigma^2_nI)^{-1}\mathbf{y_a} ./ \diag((K + \sigma^2_nI)^{-1}), \\
    \mathbf{b}& = (K + \sigma^2_nI)^{-1}\mathbf{y_b} ./ \diag((K + \sigma^2_nI)^{-1}),
\end{align}
\end{subequations}
$\mathbf{y_a} = (y_1, \dots, y_n, 0)^{\top}$ and $\mathbf{y_b} = (0, \dots, 0, 1)^{\top}$.

Using the vectors $\mathbf{a}$ and $\mathbf{b}$ from (\ref{eq:AandB1}) or (\ref{eq:AandB2}) it is now possible to 
efficiently calculate the set of output values $\tilde y$ for which $p(\tilde y) > \delta$ for any given 
significance level $\delta$. For each $i = 1, \dots, l+1$, let
\begin{align} 
	S_i &= \{\tilde y : \alpha_i(\tilde y) \geq \alpha_{l+1}(\tilde y)\}\notag\\
        &= \{\tilde y : |a_i + b_i \tilde y| \geq |a_{l+1} + b_{l+1} \tilde y|\}.\label{eq:si}
\end{align}
Each set $S_i$ (always closed) will either be an interval, a ray, the union of 
two rays, the real line, or empty; it can also be a point, which is a special
case of an interval. As we are interested in the absolute values $|a_i + b_i \tilde y|$ we can assume that
$b_i \geq 0$ for $i = 1, \dots, l+1$ (if not we multiply both $a_i$ and $b_i$ by $-1$).
If $b_i \neq b_{l+1}$, then $\alpha_i(\tilde y)$ and $\alpha_{l+1}(\tilde y)$ are 
equal at two points (which may coincide):
\begin{equation}
\label{eq:points}
	 -\frac{a_i - a_{l+1}}{b_i - b_{l+1}}\ \ \ \ {{{{{{{{{{\rm and}}}}}}}}}}\ \ \ \ -\frac{a_i + a_{l+1}}{b_i + b_{l+1}}\ ;
\end{equation}
in this case $S_i$ is an interval (maybe a point) or the union of two rays. If
$b_i = b_{l+1} \neq 0$, then $\alpha_i(\tilde y) = \alpha_{l+1}(\tilde y)$ at 
just one point:
\begin{equation}
\label{eq:point}
	 -\frac{a_i + a_{l+1}}{2b_i},
\end{equation}
and $S_i$ is a ray, unless $a_i = a_{l+1}$ in which case $S_i$ is the real line. 
If $b_i = b_{l+1} = 0$, then $S_i$ is either empty or the real line.

\begin{figure}[tbp]
    \centering
    \begin{tikzpicture}[transform shape, scale=0.9]
        \draw[latex-] (-4,0) -- (0.6,0); 
        \draw[-latex] (1.3,0) -- (4,0); 
        
        \draw[shift={(-2.5,0)},color=black] (0pt,3pt) -- (0pt,-3pt); 
        \draw[shift={(2.5,0)},color=black] (0pt,3pt) -- (0pt,-3pt); 
        
        \draw[shift={(-0.75,0)},color=black] (0pt,3pt) -- (0pt,-3pt); 
        \node at (1,0) {$\cdots$}; 
        
        \node at (-2.5,0.4) {$y_{(1)}$};
        \node at (-0.75,0.4) {$y_{(2)}$};
        \node at (2.5,0.4) {$y_{(u)}$};
        
        \draw[densely dashed] (-2.5,-0.1) -- (-2.5,-1);
        \draw[densely dashed] (-0.75,-0.1) -- (-0.75,-1);
        \draw[densely dashed] (2.5,-0.1) -- (2.5,-1);
        
        \node at (-2.5,-1.2) {$M(1)$};
        \node at (-0.75,-1.2) {$M(2)$};
        \node at (2.5,-1.2) {$M(u)$};
        
        \draw[decorate,decoration={brace,amplitude=5pt,mirror,raise=2pt},yshift=0pt]
            (-3.95,-0.1) -- (-2.55,-0.1) node [midway,below=6pt] {$N(0)$};
        \draw[decorate,decoration={brace,amplitude=5pt,mirror,raise=2pt},yshift=0pt]
            (-2.45,-0.1) -- (-0.8,-0.1) node [midway,below=6pt] {$N(1)$};
        \draw[decorate,decoration={brace,amplitude=5pt,mirror,raise=2pt},yshift=0pt]
            (2.55,-0.1) -- (3.95,-0.1) node [midway,below=6pt] {$N(u)$}; 
        
        \node at (-4,0) [left] {$-\infty$};
        \node at (4,0) [right] {$+\infty$};
    \end{tikzpicture}
    \caption{Correspondence of $N$ and $M$ to the intervals and points of $P$.}
    \label{fig:NandM}
\end{figure}
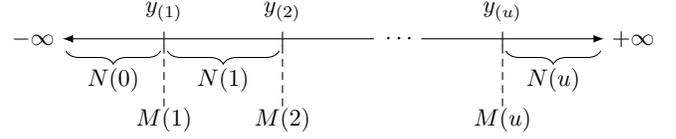

To calculate the p-value $p(\tilde y)$ for any potential output $\tilde y$ of 
the new input $x_{l+1}$, we count how many $S_i$ include $\tilde y$ and 
divide by $l+1$,
\begin{equation}
  p(\tilde y) = \frac{\#\{i = 1, \dots, l+1 : \tilde y \in S_i\}}{l+1}.
\end{equation}
As $\tilde y$ changes value, $p(\tilde y)$ can only change at the points (\ref{eq:points}) and 
(\ref{eq:point}), so for any significance level $\delta$ we can find the set of
$\tilde y$ for which $p(\tilde y) > \delta$ as the union of finitely many intervals and rays.
Algorithm~\ref{alg:alg1} implements a slightly modified version of this idea. It 
creates a list of the points (\ref{eq:points}) and (\ref{eq:point}), 
sorts it in ascending order obtaining $y_{(1)}, \dots, y_{(u)}$ and 
adds $y_{(0)} = -\infty$ to the beginning and $y_{(u+1)} = \infty$ to the end of 
this list. It then computes $N(j)$, the number of $S_i$ containing the 
interval $(y_{(j)},y_{(j+1)})$, for $j = 0, \dots, u$, and $M(j)$ the number of 
$S_i$ containing the point $y_{(j)}$, for~$j = 1, \dots, u$. Figure~\ref{fig:NandM} 
illustrates the correspondence of the elements of $N$ and $M$ to these intervals and points.

\begin{algorithm}[t]
\caption{GPR-CP.}\label{alg:alg1}
\begin{algorithmic}
    \REQUIRE training set $\{z_1, \dots, z_l\}$, new input $x_{l+1}$, covariance function $k$, vector of hyperparameters $\mathbf{\theta}$ and significance level $\delta$.
    \STATE
    \STATE Calculate $\mathbf{a}$ and $\mathbf{b}$ as defined by (\ref{eq:AandB1}) or (\ref{eq:AandB2})
    \STATE $P \leftarrow \{\}$
    \FOR{$i = 1$ to $l+1$}
        \IF{$b_i < 0$} 
            \STATE $a_i \leftarrow -a_i$
            \STATE $b_i \leftarrow -b_i$
        \ENDIF
        \IF{$b_i \neq b_{l+1}$} 
            \STATE Add (\ref{eq:points}) to $P$;
        \ELSIF{$b_i = b_{l+1} \neq 0$ \AND $a_i \neq a_{l+1}$}
            \STATE Add (\ref{eq:point}) to $P$
        \ENDIF
    \ENDFOR
    \STATE Sort $P$ in ascending order obtaining $y_{(1)}, \dots, y_{(u)}$
    \STATE Add $y_{(0)} \leftarrow -\infty$ and $y_{(u+1)} \leftarrow \infty$ to $P$
    \STATE $N(j) \leftarrow 0$, $j = 0, \dots, u$
    \STATE $M(j) \leftarrow 0$, $j = 1, \dots, u$
    \FOR{$i = 1$ to $l+1$}
				\STATE $N(j) \leftarrow N(j) + 1, j = 0, \dots, u : (y_{(j)}, y_{(j+1)}) \in S_i$
				\STATE $M(j) \leftarrow M(j) + 1, j = 1, \dots, u : y_{(j)} \in S_i$
    \ENDFOR
    \RETURN the prediction region\\
    $\mathcal{C}^{\delta}(x_{l+1}) = \left(\cup_{j:\frac{N(j)}{l+1}>\delta}\left(y_{(j)}, y_{(j+1)}\right)\right) \cup \{y_{(j)}:\frac{M(j)}{l+1}>\delta\}$.
\end{algorithmic}
\label{alg1}
\end{algorithm}

\begin{remark}
The prediction region $\mathcal{C}^{\delta}(x_{l+1})$ produced by Algorithm~\ref{alg:alg1} may not 
necessarily form a continuous interval as it may contain `holes' within it. In practice, such cases are extremely 
rare; indeed, they were not observed in any experiments conducted as part of this study. However, to ensure the 
consistent production of PIs, Algorithm~\ref{alg:alg1} can be adapted to output the convex hull of 
$\mathcal{C}^{\delta}(x_{l+1})$.
\end{remark}

\section{A Normalized Nonconformity Measure}\label{sec:nm2}

Normalized nonconformity measures~\cite{papa:jairnnr} divide the typical regression nonconformity 
with a measure of the difficulty of the particular input for the underlying algorithm. 
The intuition behind this is that two observations with the same nonconformity score as defined by (\ref{eq:nm1a}) or (\ref{eq:nm1b}), 
are not equally nonconforming if one of them is more difficult for the underlying algorithm. 
This differentiation is based on the expectation that the more difficult observation is likely to have a 
less accurate prediction, and therefore, it should be considered less nonconforming than the other, 
since their prediction accuracy is equal. Consequently, normalized nonconformity measures result 
in more precise PIs, which are tighter for inputs that are easier to predict and 
wider for inputs that are more difficult to predict.

In the case of Gaussian Process Regression, a readily available measure of the difficulty of 
an input $x_i$ is its predictive variance $\hat{\ddot{\sigma}}^2_i$. Note that this entails that $z_i$ 
is not included in the training set, corresponding to nonconformity measure definition~(\ref{eq:nm1b}) and 
the GPR leave-one-out predictive variance~(\ref{eq:gploovar}). Therefore, a normalized nonconformity 
measure for GPR-CP can be defined as
\begin{equation}
\label{eq:nm2}
	\alpha_i = \left| \frac{y_i - \hat{\ddot{y_i}}}{\sqrt[\gamma]{\hat{\ddot{\sigma}}^2_i}} \right|,
\end{equation}
where the root parameter $\gamma$ controls the sensitivity of the measure to changes of $\hat{\ddot{\sigma}}^2_i$.
The new definitions of $\mathbf{a}$ and $\mathbf{b}$ can be obtained by dividing each $a_i$ and $b_i$ 
in (\ref{eq:AandB2}) by
\begin{equation}
\sqrt[\gamma]{\frac{1}{[(K + \sigma^2_nI)^{-1}]_{ii}}},
\end{equation}
Specifically,
\begin{subequations}\label{eq:AandB3}
\begin{align}
    \mathbf{a}& = (K + \sigma^2_nI)^{-1}\mathbf{y_a} ./ \diag((K + \sigma^2_nI)^{-1})^{1 - \frac{1}{\gamma}}, \\
    \mathbf{b}& = (K + \sigma^2_nI)^{-1}\mathbf{y_b} ./ \diag((K + \sigma^2_nI)^{-1})^{1 - \frac{1}{\gamma}},
\end{align}
\end{subequations}
where $\mathbf{v}^p$ corresponds to raising each element of vector $\mathbf{v}$ to the power $p$.

Note that this nonconformity measure includes both nonconformity measures (\ref{eq:nm1a}) and (\ref{eq:nm1b}) 
as special cases when $\gamma = 1$ and $\gamma = \infty$ respectively. To see this compare the definitions of 
$\mathbf{a}$ and $\mathbf{b}$ in (\ref{eq:AandB3}) with those in (\ref{eq:AandB1}) and (\ref{eq:AandB2}). 
When $\gamma = 1$ there is a small 
difference between (\ref{eq:AandB3}) and (\ref{eq:AandB1}), namely the multiplication of all nonconformity 
scores by $\sigma^2_n$ in (\ref{eq:AandB1}). This however, does not change the resulting p-values and 
PIs since the points where 
$a_i(\tilde y) = a_{l+1}(\tilde y)$ for $i = 1, \dots, l$ remain exactly the same.

\section{Experimental Evaluation}\label{sec:exp}

This Section begins with an assessment of the performance of the proposed approach and the original GPR technique on artificial data, 
considering two scenarios: one where the exact model hyperparameters are known and one where the hyperparameters 
are unknown. Subsequently, it evaluates GPR-CP using four benchmark data sets from the UCI Machine Learning repository~\cite{data:uci}. 
This evaluation includes comparisons with the original GPR technique, four recently proposed recalibration 
techniques~\cite{kule:calib,song:distcal,marx:modular,vovk:cps} and two existing transductive CPs~\cite{nouret:tcm-rr,papa:jairnnr}. 

The implementations 
of both GPR and GPR-CP are based on the GPML Matlab code developed by Carl Edward Rasmussen and Hannes Nickisch, 
available on the website of~\cite{rasmussen:gp}. Additionally, the implementations of the four recalibration techniques 
are based on the TorchUQ~\cite{marx:modular} and Crepes~\cite{bostr:crepes} Python libraries as well as the code 
accompanying~\cite{song:distcal}.

\subsection{Evaluation Metrics}

As the aim of this work is the provision of PIs, the evaluation is centered on their calibration and tightness, 
at the confidence levels of $90\%$, $95\%$ and $99\%$. Calibration is assessed by examining 
whether the empirical miscoverage rate of the PIs matches the nominal miscoverage rate (one minus the required 
confidence level). This is in effect the 
guarantee provided by CP, see Theorem~\ref{thm:coverage}. Tightness is assessed by the mean width of the PIs. 
Narrower PIs are more precise, provided they align with the expected miscoverage rates. Significantly 
exceeding the nominal miscoverage rate marks a critical shortfall, indicating unreliable PIs, regardless of how tight they are.

It's important to highlight that recalibration techniques are commonly assessed based on their sharpness, with metrics 
such as the Negative Log Likelihood (NLL) and Continuous Ranked Probability Score (CRPS), and their calibration, using 
metrics like the Expected Calibration Error (ECE). As these metrics are not directly applicable to PIs, the comparison 
performed concentrates on the PIs derived from the outputs of recalibration techniques at the specified confidence 
levels.

\subsection{Artificial Data}

The first set of experiments compares the PIs produced by the proposed approach and the original 
GPR technique (based on the output variance for each instance) on artificially 
generated data when the model is well-specified. 

Ten data sets were generated consisting of 
$500$ training and $1000$ test observations with inputs drawn randomly from a standard Gaussian distribution. 
The outputs of each data set were generated as $y_i = f(x_i) + \epsilon$, where $\epsilon$ is Gaussian noise 
$\mathcal{N}(\epsilon|0,\sigma^2_n)$ with $\sigma_n = 0.1$ and $f$ is a Gaussian Process with a zero mean function and 
a squared exponential (SE) covariance function with covariance hyperparameters $(l, \sigma_f) = (1, 1)$; i.e. a 
unit length scale and a unit signal magnitude. The same mean function, covariance function and hyperparameters 
were used for the application of the 
two methods on these data sets. GPR-CP utilized the normalized nonconformity measure (\ref{eq:nm2}), corresponding to 
the vectors $\mathbf{a}$ and $\mathbf{b}$ defined in (\ref{eq:AandB3}), with $\gamma$ set to 
$1$, $2$ and $\infty$; recall that the first and last of these values of $\gamma$ are equivalent to definitions 
(\ref{eq:nm1a}) and (\ref{eq:nm1b}) respectively. 

\begin{table}[t]
  \centering
  \begin{tabular}{lcrrrrrr} \hline\noalign{\smallskip}
                          \Lower{Method} & \Lower{$\gamma$} & \multicolumn{3}{c}{Mean PI Width} 
													               & \multicolumn{3}{c}{Miscoverage (\%)}\\
                                         & & \multicolumn{1}{c}{90\%} & \multicolumn{1}{c}{95\%}
                                         & \multicolumn{1}{c}{99\%} 
                                         & \multicolumn{1}{c}{90\%} & \multicolumn{1}{c}{95\%}
                                         & \multicolumn{1}{c}{99\%}\\ \noalign{\smallskip}\hline\noalign{\smallskip}

    GPR       &          & 1.219 & 1.453 & 1.909 & 10.27 & 4.90 & 1.02\\ \noalign{\smallskip}\hline \noalign{\smallskip}
              & $1$      & 2.360 & 3.167 & 4.823 & 10.60 & 4.73 & 0.91\\
    GPR-CP    & $2$      & \textbf{\underline{1.211}} & \textbf{\underline{1.442}} & \textbf{\underline{1.905}} & 10.51 & 5.14 & 1.09\\
              & $\infty$ & 1.377 & 1.853 & 2.953 &  9.88 & 5.04 & 1.06\\ \noalign{\smallskip}\hline \noalign{\smallskip}
  \end{tabular}
  \caption{Comparison of the proposed approach and the original GPR on artificial data when the model is well-specified (the smallest PI widths appear in bold and underlined).}
  \label{tab:artificial1}
\end{table}

\begin{table}
  \centering
  \begin{tabular}{lcrrrrrr} \hline\noalign{\smallskip}
                          \Lower{Method} & \Lower{$\gamma$} & \multicolumn{3}{c}{Mean PI Width} 
													               & \multicolumn{3}{c}{Miscoverage (\%)}\\
                                         & & \multicolumn{1}{c}{90\%} & \multicolumn{1}{c}{95\%}
                                         & \multicolumn{1}{c}{99\%} 
                                         & \multicolumn{1}{c}{90\%} & \multicolumn{1}{c}{95\%}
                                         & \multicolumn{1}{c}{99\%}\\ \noalign{\smallskip}\hline\noalign{\smallskip}

    GPR       &          & 1.834 & 2.185 & 2.872 &  8.51 & 5.21 & 2.59\\ \noalign{\smallskip}\hline \noalign{\smallskip}
              & $1$      & 1.935 & 2.595 & 5.162 & 10.25 & 4.74 & 1.00\\
    GPR-CP    & $2$      & \textbf{\underline{1.716}} & \textbf{\underline{2.259}} & \textbf{\underline{3.866}} & 10.29 & 4.78 & 1.03\\
              & $\infty$ & 1.875 & 2.532 & 4.092 & 10.45 & 5.09 & 0.99\\ \noalign{\smallskip}\hline \noalign{\smallskip}
  \end{tabular}
  \caption{Comparison of the proposed approach and the original GPR on artificial data with outliers and unknown hyperparameters (the smallest PI widths of GPR-CP appear in bold and underlined).}
  \label{tab:artificial2}
\end{table}

Table~\ref{tab:artificial1} reports the mean width of the PIs produced by 
each method for all ten data sets together with the corresponding miscoverage percentages. 
In this case the PIs produced by both methods are well-calibrated, as their miscoverage rates 
are very close to the required significance levels. 
In terms of PI tightness, the 
smallest widths of GPR-CP are obtained with the normalized version of the nonconformity measure (i.e. $\gamma = 2$) 
and are slightly smaller than those obtained by the original GPR technique.

A second set of experiments was performed on the same type of data, however this time with probability $0.1$ ``outlier'' 
observations were generated, for which the noise standard deviation was $1$ instead of $0.1$. Again ten 
data sets were generated consisting of $500$ training and $1000$ test observations. Additionally, the zero mean function and 
SE covariance function that were used for generating the data were also used for the application of the two methods on 
these data sets. In this case though, the hyperparameters (length scale, signal magnitute and noise standard deviation) 
were optimized by minimizing the negative log marginal likelihood.

Table~\ref{tab:artificial2} reports the results of the proposed approach and the original GPR technique on these 
data sets. Although the miscoverage rates of the PIs produced by the original GPR 
technique for the $90\%$ and $95\%$ 
confidence levels are again near or below the required significance levels, that of the $99\%$ 
confidence level is more than double the required level. This demonstrates experimentally that 
when the GPR model is not well-specified, which is typically the case for real data, 
the resulting PIs become misleading. On the contrary, the corresponding miscoverage rates of 
the proposed approach are again in all cases very close to the required levels. In terms of PI 
tightness, the normalized version of the nonconformity measure ($\gamma = 2$) again gave the smallest widths of 
GPR-CP, which are smaller than the ones produced by the original GPR technique 
for the $90\%$ confidence level and quite close to those produced by the original GPR for the $95\%$ confidence level.

\subsection{Benchmark Data}

\begin{table*}[t]
  \centering
  \begin{tabular}{lccc@{\hspace{2mm}}ccccccc@{\hspace{2mm}}cccccc} \hline \rule{0pt}{10pt}
	                                       & & & & \multicolumn{6}{c}{Boston Housing} & & \multicolumn{6}{c}{Auto-mpg}\\[2pt]
	\hline \rule{0pt}{10pt}
                          \Lower{Method} & Covariance & \Lower{$\gamma$} & & \multicolumn{3}{c}{Mean PI Width} 
													               & \multicolumn{3}{c}{Miscoverage (\%)} 
                                         & & \multicolumn{3}{c}{Mean PI Width} & \multicolumn{3}{c}{Miscoverage (\%)} \\
                                         & Function & & & \multicolumn{1}{c}{90\%} & \multicolumn{1}{c}{95\%}
                                         & \multicolumn{1}{c}{99\%} 
                                         & \multicolumn{1}{c}{90\%} & \multicolumn{1}{c}{95\%}
                                         & \multicolumn{1}{c}{99\%} & & \multicolumn{1}{c}{90\%} & \multicolumn{1}{c}{95\%}
                                         & \multicolumn{1}{c}{99\%} 
                                         & \multicolumn{1}{c}{90\%} & \multicolumn{1}{c}{95\%}
                                         & \multicolumn{1}{c}{99\%}\\ \noalign{\smallskip}\hline\noalign{\smallskip}

              & SE             &          & &  9.230 & 10.997 & 14.453   & 8.02 & 5.02 & 2.69 & &  8.307 &  9.898 & 13.009    &  8.49 & 5.28 & 2.68\\ 
              & RQ             &          & &  9.146 & 10.897 & 14.322   & 8.14 & 5.32 & 2.79 & &  8.299 &  9.888 & 12.995    &  8.55 & 5.36 & 2.70\\ 
    GPR       & NN             &          & &  9.031 & 10.760 & 14.142   & 8.66 & 5.40 & 2.67 & &  8.276 &  9.861 & 12.960    &  8.57 & 5.28 & 2.70\\ 
              & Matern $(3/2)$ &          & &  9.093 & 10.834 & 14.239   & 7.83 & 5.38 & 2.87 & &  8.297 &  9.886 & 12.993    &  8.67 & 5.48 & 2.60\\ 
              & Matern $(5/2)$ &          & &  9.116 & 10.862 & 14.276   & 8.08 & 5.26 & 2.77 & &  8.298 &  9.887 & 12.994    &  8.65 & 5.41 & 2.68\\ \noalign{\smallskip}\hline\noalign{\smallskip}
    GPR-RI    & SE             &          & &  9.369 & 12.451 & 20.547   & 11.64 & 5.51 & 1.78 & & 8.454 & 11.506 & 19.420    & 11.12 & 6.35 & 1.71\\
    GPR-RB    & SE             &          & &  9.137 & 11.469 & 16.788   & 11.01 & 6.62 & 2.41 & & 8.134 & 10.208 & 14.898    & 11.71 & 6.53 & 1.86\\
		GPR-RM    & SE             &          & &  9.457 & 12.133 & 20.471   & 11.17 & 5.63 & 2.17 & & 8.489 & 10.954 & 18.440    & 10.99 & 5.79 & 1.86\\
		GPR-RC * & SE             &          & & 10.372 & 16.205 & $\infty$ &  9.01 & 3.52 & 0.00  & & 9.710 & 13.188 & $\infty$ & 8.93 & 4.54 & 0.00\\ \noalign{\smallskip}\hline\noalign{\smallskip}
    RR-CP *     & SE             &          & & 11.651 & 15.042 & 27.074 & 9.86 & 4.72 & 0.83 & & 8.015 & 10.235 & 20.696 & 9.80 & 4.92 & 0.84\\
    $k$NNR-CP * &                &          & & 11.592 & 15.300 & 25.721 & 9.76 & 4.90 & 0.89 & & 8.521 & 11.587 & 21.383 & 9.57 & 4.90 & 0.84\\	\noalign{\smallskip}\hline\noalign{\smallskip}
              &                & $1$      & & 9.678 & 12.978 & 23.197 & 10.36 & 4.84 & 0.87 & & 7.801 & 10.302 & 19.321 & 10.36 & 4.87 & 0.82\\
              &                & $2$      & & 8.277 & 11.078 & 19.773 & 10.53 & 4.94 & 0.91 & & 7.746 & 10.286 & 19.315 & 10.20 & 4.87 & 0.82\\
\Lower{GPR-CP *} & \Lower{SE}    & $3$      & & \textbf{\underline{8.140}} & 10.671 & \textbf{\underline{19.106}} & 10.87 & 4.98 & 1.13 & & \textbf{\underline{7.737}} & 10.300 & 19.349 & 10.10 & 4.97 & 0.82\\
							&                & $4$      & & 8.149 & \textbf{\underline{10.589}} & 19.978 & 11.23 & 5.16 & 0.95 & & 7.745 & 10.303 & 19.379 & 10.18 & 4.97 & 0.82\\
							&                & $8$      & & 8.282 & 10.659 & 20.584 & 11.25 & 5.47 & 1.03 & & 7.761 & 10.315 & 19.471 & 10.23 & 4.97 & 0.87\\
              &                & $\infty$ & & 8.468 & 10.938 & 22.192 & 11.03 & 5.51 & 0.91 & & 7.777 & 10.341 & 19.687 & 10.23 & 5.00 & 0.87\\ \noalign{\smallskip}\hline\noalign{\smallskip}
              &                & $1$      & & 9.878 & 13.391 & 24.248 & 10.34 & 4.94 & 0.97 & & 7.812 & 10.275 & 19.256 & 10.46 & 5.08 & 0.84\\
    GPR-CP *   & RQ             & $2$      & & 8.212 & 11.197 & 20.147 & 10.53 & 4.96 & 1.03 & & 7.749 & 10.273 & 19.270 & 10.31 & 5.03 & 0.87\\
              &                & $\infty$ & & 8.368 & 10.845 & 23.501 & 11.01 & 5.55 & 0.95 & & 7.778 & 10.369 & 19.685 & 10.33 & 5.03 & 0.87\\ \noalign{\smallskip}\hline\noalign{\smallskip}
              &                & $1$      & & 9.528 & 12.863 & 21.461 & 10.22 & 4.80 & 0.95 & & 7.794 & 10.049 & 18.984 & 10.41 & 5.13 & 0.82\\
    GPR-CP *   & NN             & $2$      & & 8.273 & 11.029 & 19.533 & 10.61 & 5.10 & 0.91 & & 7.769 & \textbf{\underline{10.013}} & \textbf{\underline{18.920}} & 10.46 & 5.08 & 0.82\\
              &                & $\infty$ & & 8.349 & 11.031 & 22.270 & 10.51 & 5.18 & 0.91 & & 7.855 & 10.080 & 19.063 & 10.56 & 5.03 & 0.92\\ \noalign{\smallskip}\hline\noalign{\smallskip}
              &                & $1$      & & 9.797 & 13.474 & 24.718 & 10.18 & 4.84 & 0.87 & & 7.874 & 10.283 & 18.932 & 10.13 & 5.08 & 0.82\\
    GPR-CP *   & Matern $(3/2)$ & $2$      & & 8.144 & 11.306 & 20.239 & 10.32 & 4.86 & 1.01 & & 7.764 & 10.374 & 19.017 & 10.33 & 5.03 & 0.82\\
              &                & $\infty$ & & 8.247 & 10.712 & 24.463 & 10.89 & 5.34 & 0.87 & & 7.793 & 10.565 & 19.681 & 10.28 & 5.03 & 0.87\\ \noalign{\smallskip}\hline\noalign{\smallskip}
              &                & $1$      & & 9.858 & 13.375 & 24.173 & 10.20 & 4.82 & 0.95 & & 7.880 & 10.245 & 19.100 & 10.15 & 5.10 & 0.82\\
    GPR-CP *   & Matern $(5/2)$ & $2$      & & 8.159 & 11.206 & 20.040 & 10.45 & 4.88 & 1.03 & & 7.770 & 10.313 & 19.156 & 10.28 & 5.08 & 0.82\\
              &                & $\infty$ & & 8.342 & 10.820 & 23.506 & 10.99 & 5.45 & 0.95 & & 7.767 & 10.440 & 19.708 & 10.41 & 4.92 & 0.87\\ \noalign{\smallskip}\hline\noalign{\smallskip}
  \end{tabular}
  \caption{Results of all approaches on the Boston Housing and Auto-mpg data sets (* indicates the approaches that produce well-calibrated PIs across all confidence levels; the best mean PI widths over well-calibrated approaches appear in bold and underlined).}
  \label{tab:bostonmpg}
\end{table*}

\begin{table*}[t]
  \centering
  \begin{tabular}{lccc@{\hspace{2mm}}ccccccc@{\hspace{2mm}}cccccc} \hline \rule{0pt}{10pt}
	                                       & & & & \multicolumn{6}{c}{CPU Performance} & & \multicolumn{6}{c}{Servo}\\[2pt]
	\hline \rule{0pt}{10pt}
                          \Lower{Method} & Covariance & \Lower{$\gamma$} & & \multicolumn{3}{c}{Mean PI Width} 
													               & \multicolumn{3}{c}{Miscoverage (\%)} 
                                         & & \multicolumn{3}{c}{Mean PI Width} & \multicolumn{3}{c}{Miscoverage (\%)} \\
                                         & Function & & & \multicolumn{1}{c}{90\%} & \multicolumn{1}{c}{95\%}
                                         & \multicolumn{1}{c}{99\%} 
                                         & \multicolumn{1}{c}{90\%} & \multicolumn{1}{c}{95\%}
                                         & \multicolumn{1}{c}{99\%} & & \multicolumn{1}{c}{90\%} & \multicolumn{1}{c}{95\%}
                                         & \multicolumn{1}{c}{99\%} 
                                         & \multicolumn{1}{c}{90\%} & \multicolumn{1}{c}{95\%}
                                         & \multicolumn{1}{c}{99\%}\\ \noalign{\smallskip}\hline\noalign{\smallskip}

              & SE             &          & & 108.78 & 129.61 & 170.34 & 12.58 & 9.33 & 5.55 & & 1.853 & 2.208 & 2.902 &  9.04 & 7.01 & 4.67\\
              & RQ             &          & & 111.12 & 132.40 & 174.02 & 12.15 & 9.14 & 5.55 & & 1.823 & 2.172 & 2.854 &  9.82 & 7.72 & 4.67\\
    GPR       & NN             &          & & 100.52 & 119.77 & 157.41 & 13.54 & 9.43 & 5.41 & & 1.951 & 2.324 & 3.054 &  7.90 & 6.23 & 3.83\\
              & Matern $(3/2)$ &          & & 108.34 & 129.09 & 169.66 & 11.53 & 8.66 & 5.45 & & 1.801 & 2.146 & 2.821 &  8.74 & 7.07 & 4.91\\
              & Matern $(5/2)$ &          & & 107.75 & 128.39 & 168.74 & 12.01 & 8.61 & 5.60 & & 1.810 & 2.157 & 2.835 &  9.34 & 7.01 & 4.97\\ \noalign{\smallskip}\hline\noalign{\smallskip}
    GPR-RI    & SE             &          & & 133.93 & 182.25 & 285.59 & 11.87 & 6.79 & 2.25 & & 1.919 & 2.772 & 4.383 & 11.38 & 6.95 & 3.59\\
    GPR-RB    & SE             &          & & 104.65 & 131.84 & $\infty$ & 13.35 & 8.04 & 3.16 & & 1.348 & $\infty$ & $\infty$ & 9.22 & 5.81 & 2.93\\
    GPR-RM    & SE             &          & & 137.75 & 191.82 & 279.07 & 12.39 & 7.75 & 3.06 & & 1.902 & 2.639 & 4.133 & 11.14 & 7.01 & 3.77\\
    GPR-RC *   & SE             &          & & 165.19 & 228.18 & $\infty$ & 9.67 & 4.55 & 0.00 & & 3.701 & $\infty$ & $\infty$ & 4.73 & 0.00 & 0.00\\	\noalign{\smallskip}\hline\noalign{\smallskip}
    RR-CP *     & SE             &          & & $\infty$ & $\infty$ & $\infty$ & 9.90 & 4.64 & 0.57 & & 1.869 & 2.945 & 7.224 &  9.76 & 4.31 & 0.60\\
    $k$NNR-CP * &                &          & & 141.57   & 206.03   & 545.31   & 9.38 & 4.55 & 0.62 & & \textbf{\underline{1.423}} & 3.023 & 7.675 & 10.42 & 4.55 & 0.60\\	\noalign{\smallskip}\hline\noalign{\smallskip}
              &                & $1$      & & 160.57 & 221.37 & 333.62 & 10.57 & 5.79 & 1.05 & & 1.731 & 2.989 & 6.171 & 10.36 & 4.55 & 0.54\\
              &                & $2$      & & 110.33 & 165.04 & \textbf{\underline{253.38}} & 10.96 & 5.17 & 1.05 & & 1.650 & 2.872 & 6.699 & 10.24 & 4.55 & 0.66\\
\Lower{GPR-CP *} & \Lower{SE}    & $3$      & & \textbf{\underline{106.65}} & 157.14 & 295.21 & 11.77 & 5.89 & 0.77 & & 1.631 & 2.846 & 6.908 & 10.36 & 4.55 & 0.66\\
							&                & $4$      & & 109.79 & \textbf{\underline{154.14}} & 332.34 & 11.48 & 6.08 & 0.91 & & 1.623 & 2.833 & 7.018 & 10.36 & 4.55 & 0.66\\
							&                & $8$      & & 115.18 & 158.23 & 414.94 & 10.91 & 6.08 & 1.15 & & 1.614 & 2.814 & 7.191 & 10.42 & 4.49 & 0.66\\
              &                & $\infty$ & & 121.52 & 176.86 & 537.25 & 10.57 & 5.55 & 1.24 & & 1.609 & 2.798 & 7.372 & 10.36 & 4.43 & 0.66\\ \noalign{\smallskip}\hline\noalign{\smallskip}
              &                & $1$      & & 161.34 & 228.68 & 389.69 & 10.86 & 5.55 & 1.05 & & 1.720 & 3.001 & 6.018 & 11.02 & 4.79 & 0.78\\
    GPR-CP *   & RQ             & $2$      & & 115.72 & 177.96 & 316.22 & 11.05 & 5.12 & 0.86 & & 1.614 & 2.875 & 6.489 & 11.14 & 4.55 & 0.66\\
              &                & $\infty$ & & 125.71 & 188.65 & 551.74 & 11.10 & 5.45 & 1.20 & & 1.574 & 2.782 & 7.129 & 11.02 & 4.67 & 0.66\\
 \noalign{\smallskip}\hline\noalign{\smallskip}
              &                & $1$      & & 139.70 & 191.58 & $\infty$ & 10.77 & 5.98 & 0.91 & & 1.688 & 2.909 & 7.314 &  9.70 & 4.25 & 0.66\\
    GPR-CP *   & NN             & $2$      & & 109.35 & 156.70 &   277.73 & 10.91 & 5.41 & 1.10 & & 1.650 & 2.835 & 7.628 &  9.76 & 4.31 & 0.60\\
              &                & $\infty$ & & 116.16 & 164.20 & $\infty$ & 11.29 & 5.31 & 0.77 & & 1.625 & \textbf{\underline{2.772}} & 7.997 &  9.88 & 4.37 & 0.60\\ \noalign{\smallskip}\hline\noalign{\smallskip}
              &                & $1$      & & 148.93 & 222.75 & 392.61 & 10.43 & 5.26 & 0.81 & & 1.774 & 3.109 & \textbf{\underline{5.832}} &  9.82 & 4.55 & 0.72\\
    GPR-CP *   & Matern $(3/2)$ & $2$      & & 111.93 & 180.24 & 316.88 & 10.57 & 4.78 & 0.77 & & 1.632 & 2.951 & 6.280 &  9.82 & 4.37 & 0.54\\
              &                & $\infty$ & & 124.25 & 189.81 & 523.97 & 10.57 & 5.26 & 0.96 & & 1.550 & 2.835 & 6.915 &  9.64 & 4.61 & 0.60\\
\noalign{\smallskip}\hline\noalign{\smallskip}
              &                & $1$      & & 160.58 & 228.93 & 386.71 & 10.48 & 5.50 & 0.91 & & 1.769 & 3.062 & 5.907 & 10.06 & 4.55 & 0.78\\
    GPR-CP *   & Matern $(5/2)$ & $2$      & & 112.22 & 177.38 & 305.51 & 10.81 & 5.22 & 0.91 & & 1.650 & 2.920 & 6.376 & 10.12 & 4.37 & 0.60\\
              &                & $\infty$ & & 124.00 & 188.27 & 532.02 & 10.81 & 5.41 & 1.10 & & 1.569 & 2.817 & 7.021 & 10.00 & 4.61 & 0.66\\ \noalign{\smallskip}\hline\noalign{\smallskip}
  \end{tabular}
  \caption{Results of all approaches on the CPU Performance and Servo data sets (* indicates the approaches that produce well-calibrated PIs across all confidence levels; the best mean PI widths over well-calibrated approaches appear in bold and underlined).}
  \label{tab:cpuservo}
\end{table*}

The performance of the proposed approach was evaluated on four benchmark data sets from the UCI Machine Learning 
repository~\cite{data:uci}:
\begin{itemize}
\item \emph{Boston Housing}, which lists the median house prices for $506$ 
	different areas of Boston MA in \$1000s. Each area is described by 
	13 inputs such as pollution and crime rate.
\item \emph{Auto-mpg}, which concerns fuel consumption in miles per gallon (mpg).
	The original data set consisted of 398 observations out of which 6 were missing input 
	values and were discarded. Furthermore, one input was also discarded as it was 
	a unique identifying string. The remaining data set had 7 inputs such as engine 
	capacity and number of cylinders.
\item \emph{CPU Performance} (or Machine), which lists the relative performance of
	209 CPUs based on six continuous inputs and 1 multi-valued discrete input (the vendor).
\item \emph{Servo}, which is is concerned with the rise time of a servo mechanism in 
    terms of two continuous gain settings and two discrete choices of mechanical linkages.
    It consists of 167 observations.
\end{itemize}

All experiments consisted of 10 random runs of a 10-fold cross-validation process for each data set. 
The inputs were normalized by setting their mean to $0$ and their standard deviation to $1$ based 
on the training examples of each fold. 

The proposed approach and original GPR technique were extensively evaluated using 5 covariance functions: Squared 
Exponential (SE), Rational Quadratic (RQ), Neural Network (NN) and Matern with smoothness set to $v = 3/2$ and $v = 5/2$. 
In all cases, hyperparameters were optimized within each training set by minimizing the negative log marginal 
likelihood. To mitigate the risk of converging to local minima, the minimization process was performed 3 times with 
different random initial values and the hyperparameters that resulted in the lowest negative log marginal likelihood were selected. 
It's worth noting that exactly the same hyperparameters were used for both methods.

The covariance functions that gave the smallest mean negative log marginal 
likelihood over all runs and folds for each data set were: Matern with $v = 3/2$ for 
Boston Housing, Rational Quadratic for Auto-mpg, Neural Network for CPU Performance, 
and Matern with $v = 5/2$ for Servo.

GPR-CP utilized the normalized nonconformity measure (\ref{eq:nm2}), corresponding to the vectors
$\mathbf{a}$ and $\mathbf{b}$ defined in (\ref{eq:AandB3}). The nonconformity measure parameter 
$\gamma$ was set to $1$, $2$ and $\infty$ in all experiments. Additionally, to further explore its impact 
on the results it was also set to $3$, $4$ and $8$ when utilizing the SE covariance function.

For comparison with previously proposed techniques, the same experiments were conducted with four recalibration 
approaches and two existing transductive CP approaches. 
All four recalibration approaches employed GPR with the SE covariance function 
as underlying technique, utilizing the Scikit-learn~\cite{scikit-learn} implementation of GPR. The number of 
instances withheld for the calibration process was set to $25\%$ of the training set (following~\cite{marx:modular}). 
It's worth noting that not employing a separate calibration set, as suggested in~\cite{kule:calib} and~\cite{song:distcal}, 
resulted in extremely uncalibrated PIs and was therefore not considered as an option.

The specific recalibration approaches considered are:
\begin{itemize}
\item Isotonic Calibration (GPR-RI): Algorithm 1 in~\cite{kule:calib} with isotonic regression as recalibrator.
\item GP-BETA Calibration (GPR-RB): Algorithm proposed in~\cite{song:distcal} with the number of inducing points set to 64.
\item Modular Conformal Calibration (GPR-RM): Algorithm 1 in~\cite{marx:modular} with the Neural Autoregressive Flow (NAF) interpolation algorithm.
\item Split Conformal Predictive System (GPR-RC): Algorithm 1 in~\cite{vovk:cps}.
\end{itemize}

The same experiments were conducted with two existing transductive CPs: the Ridge Regression CP (RR-CP)~\cite{nouret:tcm-rr} 
and the $k$-Nearest Neighbours Regression CP ($k$NNR-CP)~\cite{papa:jairnnr}. As mentioned in 
Section~\ref{sec:RW}, RR-CP uses the standard (non-normalized) regression nonconformity measure, while $k$NNR-CP is the 
only existing TCP approach that employs a normalized measure. 

RR-CP utilized an RBF kernel, which corresponds to the SE covariance 
function of GPR. The kernel parameter and ridge factor were optimized by minimizing
the leave-one-out error on each fold using the gradient descent approach proposed in~\cite{chapelle:svmparam} 
and the corresponding code provided online\footnote{The code is located at http://olivier.chapelle.cc/ams/}. 
The parameter $k$ of nearest neighbours for $k$NNR-CP was selected by performing a 10-fold cross-valdation 
process with the original $k$-NNR technique on the whole of each data set and selecting the value of $k$ that gave 
the smallest error. The use of the whole data set involves snooping, but this is in favour of $k$NNR-CP.

\begin{figure*}[pt]
  \centering
    \includegraphics[width=10.5cm, trim=15 15 15 5]{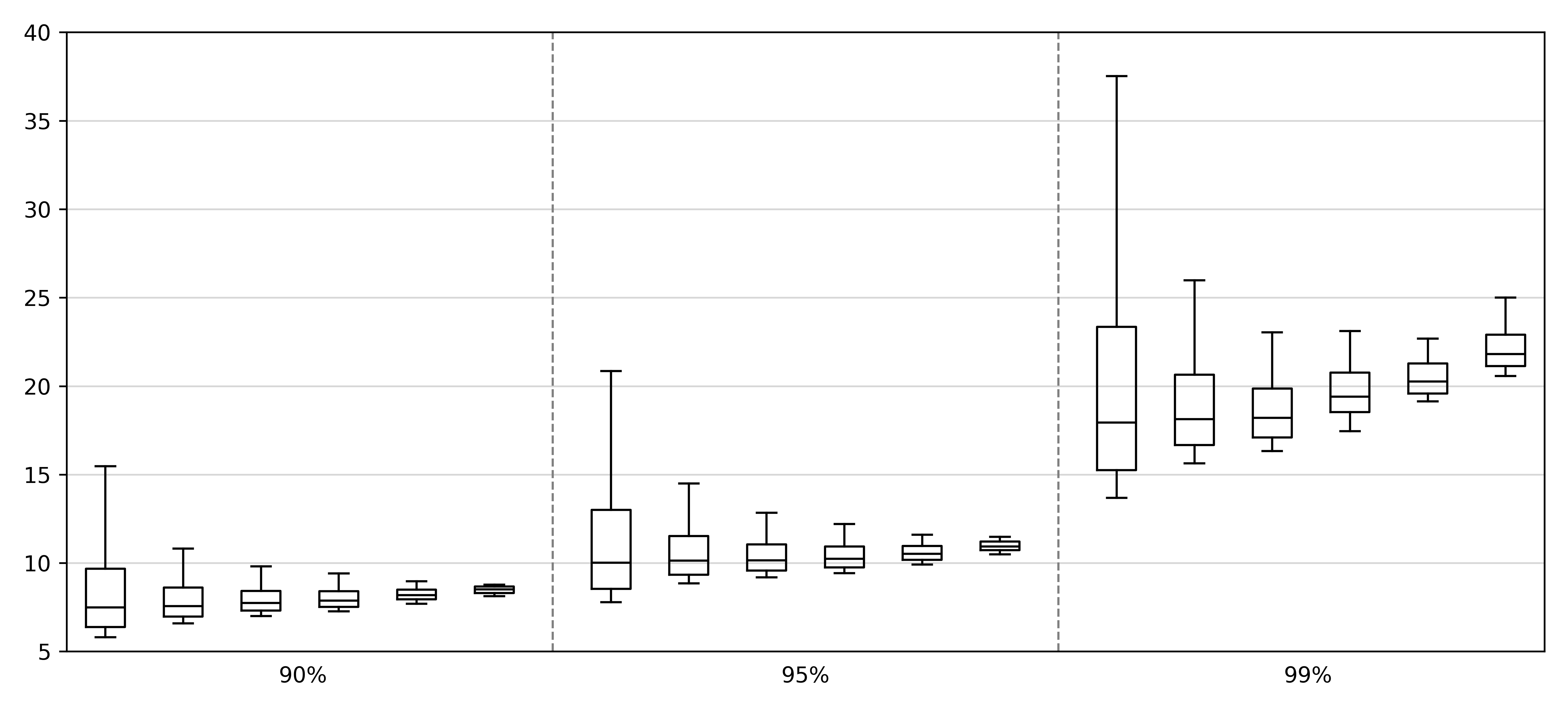}
  \caption{Boston Housing data set PI width distribution for the $90\%$, $95\%$ and $99\%$ confidence levels (Each part contains the boxplots for the PI widths produced with $\gamma$ set to $1$, $2$, $3$, $4$, $8$ and $\infty$ from left to right).}
  \label{fig:boston}
\end{figure*}

\begin{figure*}[pt]
  \centering
    \includegraphics[width=10.5cm, trim=15 15 15 5]{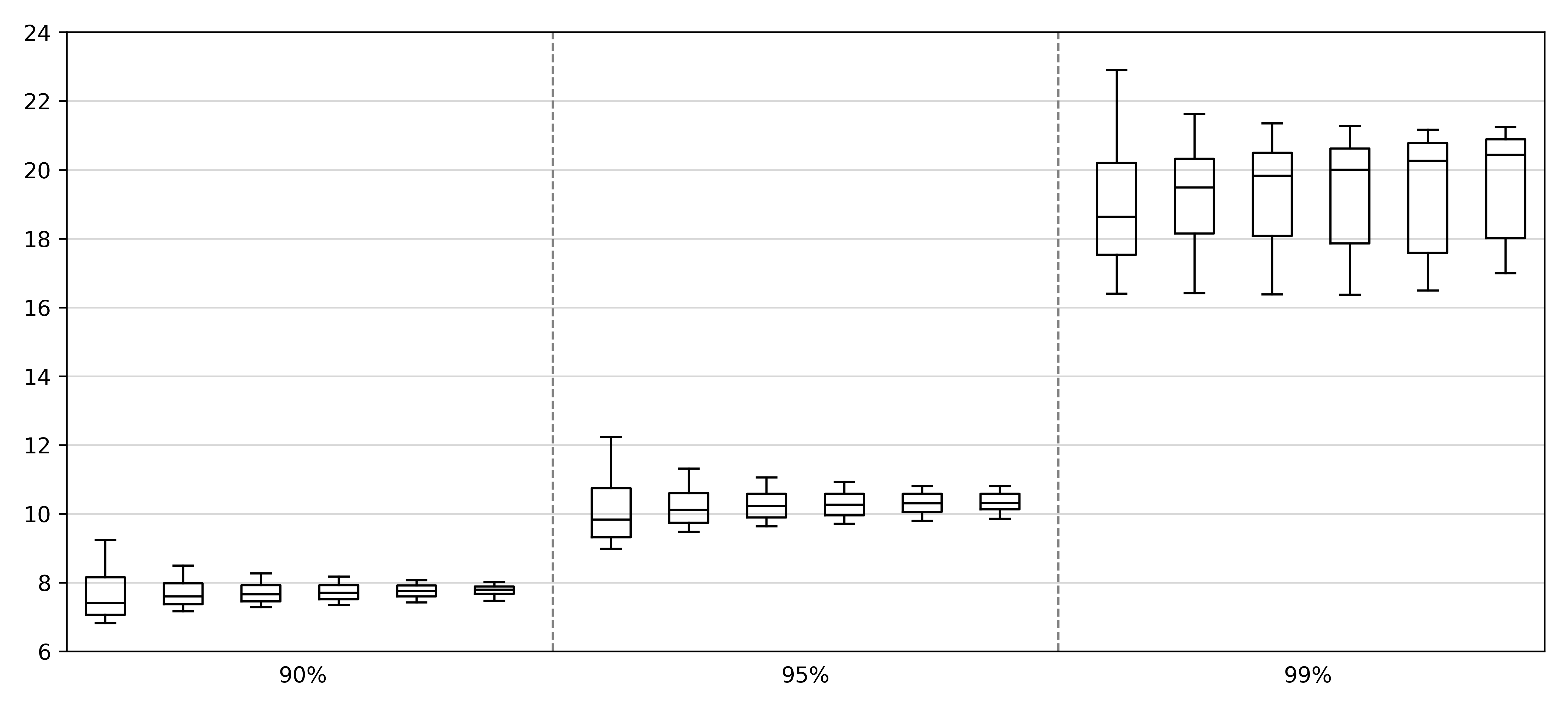}
  \caption{Auto-mpg data set PI width distribution for the $90\%$, $95\%$ and $99\%$ confidence levels (Each part contains the boxplots for the PI widths produced with $\gamma$ set to $1$, $2$, $3$, $4$, $8$ and $\infty$ from left to right).}
  \label{fig:auto-mpg}
\end{figure*}

\begin{figure*}[pt]
  \centering
    \includegraphics[width=10.5cm, trim=15 15 15 5]{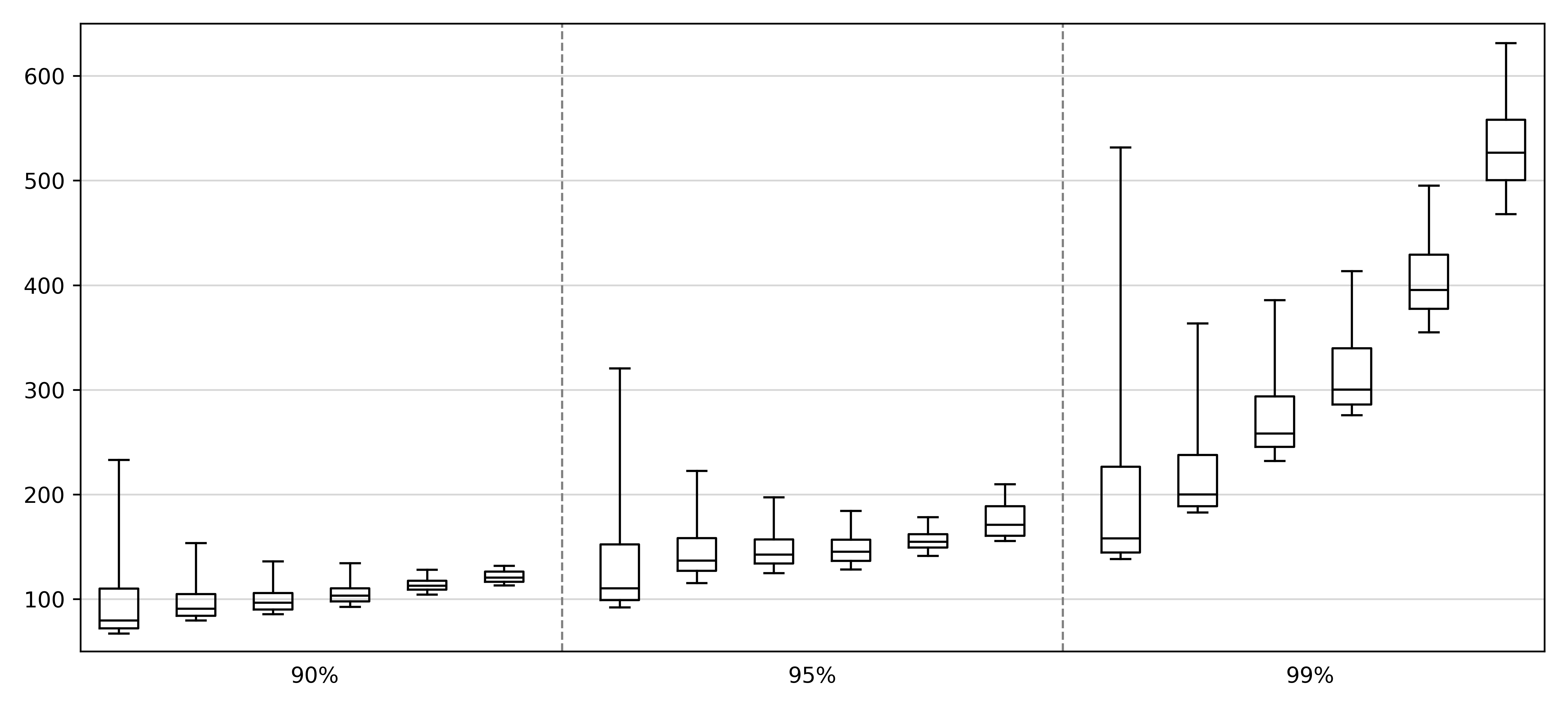}
  \caption{CPU Performance data set PI width distribution for the $90\%$, $95\%$ and $99\%$ confidence levels (Each part contains the boxplots for the PI widths produced with $\gamma$ set to $1$, $2$, $3$, $4$, $8$ and $\infty$ from left to right).}
  \label{fig:machine}
\end{figure*}

\begin{figure*}[pt]
  \centering
    \includegraphics[width=10.5cm, trim=15 15 15 5]{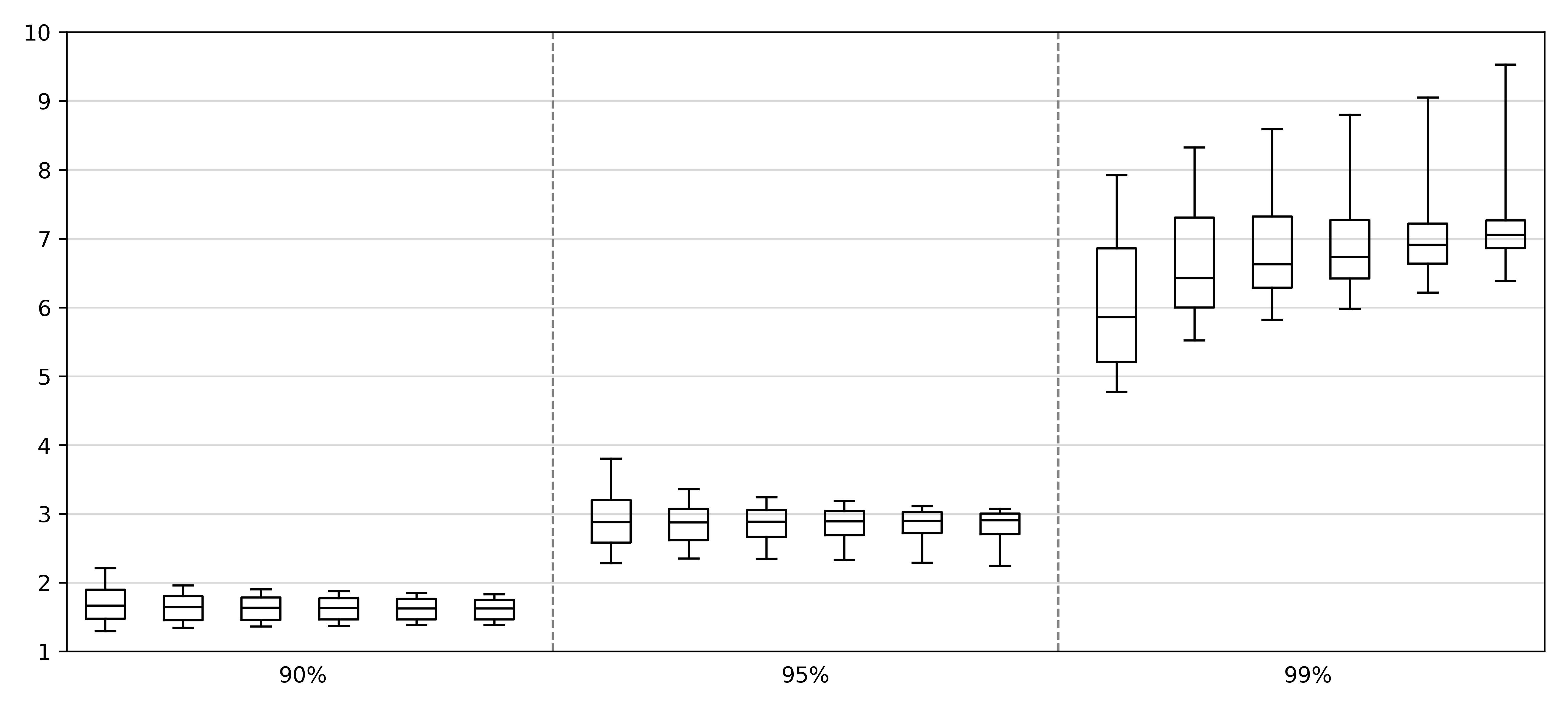}
  \caption{Servo data set PI width distribution for the $90\%$, $95\%$ and $99\%$ confidence levels (Each part contains the boxplots for the PI widths produced with $\gamma$ set to $1$, $2$, $3$, $4$, $8$ and $\infty$ from left to right).}
  \label{fig:servo}
\end{figure*}

Tables~\ref{tab:bostonmpg} and~\ref{tab:cpuservo} present the results of all approaches 
on the Boston Housing and Auto-mpg, and on the CPU Performance and Servo data sets, respectively.
Specifically, these tables present the mean PI widths together with the miscoverage percentage 
of the corresponding PIs for each confidence level.
For ease of presentation, the approaches are divided into 8 groups consisting of the original GPR with the 5 covariance 
functions, the four recalibration approaches, the two existing TCP approaches, and one group for the 
proposed approach with each of the 5 covariance functions.

By assessing the well-calibratedness of approaches one can observe that the PIs of the original GPR, GPR-RI, GPR-RB and 
GPR-RM approaches can become very misleading. Across all data sets, these approaches fail to cover the true label in significantly 
more than the required $1\%$ of cases at the $99\%$ confidence level. Moreover, for the CPU Performance and Servo data sets, 
the considerably higher than the nominal miscoverage percentages extend to the $95\%$ confidence level, and in some cases, even 
to the $90\%$ confidence level. On the contrary, the miscoverage rates of the PIs produced by the proposed approach as well as 
GPR-RC, RR-CP and $k$NNR-CP are close to the required significance levels in all cases. It is important to note here that the 
well-calibratedness of the proposed approach is not affected by the choice of covariance function or nonconformity measure.

In assessing PI tightness, the best mean PI widths among the four well-calibrated approaches are marked in bold and underlined. 
In nearly all cases GPR-CP produces the tightest mean PI widths. The sole exception is observed at the $90\%$ confidence level 
for the Servo data set. 

When comparing GPR-CP with SE covariance and $\gamma=2$ to the four recalibration approaches, one can notice that in virtually 
all cases, either the mean PI widths of GPR-CP are tighter, or the corresponding PIs have an extremely high miscoverage 
percentage, surpassing $30\%$ above the significance level. Once again the only exception is at the $90\%$ confidence level 
for the Servo data set. 

Performing the same comparison with the two existing TCP approaches shows that the mean PI 
widths of GPR-CP are again almost always better. There are only two exceptions: RR-CP gives slightly smaller widths for the 
$95\%$ confidence level on the Auto-mpg data set, and $k$NNR-CP gives smaller widths for the $90\%$ confidence level on the 
Servo data set. Another important observation in these results is the sensitivity of the non-normalized nonconformity measure 
of RR-CP, which can result in some 
extremely wide intervals. This is particularly evident in the widths produced for the CPU Performance data set, underlining 
the significant advantage of normalized nonconformity measures. 

In fact, the substantial difference that can be observed between the well-calibrated mean widths of all existing (recalibration and CP) approaches and those of GPR-CP on the Boston Housing and CPU Performance data sets demonstrates the superiority of GPR-CP 
when combined with the normalized nonconformity measure proposed in this work.

Figures~\ref{fig:boston}-\ref{fig:servo} complement the information in 
Tables~\ref{tab:bostonmpg} and \ref{tab:cpuservo} by displaying boxplots of the PI widths generated by GPR-CP with 
SE covariance for each data set. In particular, each boxplot shows the median, upper and lower 
quartiles, and upper and lower deciles of the corresponding PI widths. 
Each plot is divided into three sections corresponding to the three confidence levels of interest ($90\%$, $95\%$ and $99\%$) 
and each section contains the boxplots for the PI widths produced with $\gamma$ set to $1$, 
$2$, $3$, $4$, $8$, and $\infty$ arranged from left to right. 

A comparison of the results obtained with the different values of $\gamma$, firstly demonstrates the substantial positive 
impact of the normalized nonconformity measure on the tightness of the resulting PIs. This 
is evident from the PIs generated when $\gamma$ is set to $\infty$, which disregards the variance of observations. 
These PIs have the largest median widths across almost all cases. Notably, for the Boston Housing and CPU Performance data sets 
the large majority of the PIs produced with $\gamma = \infty$ are wider than the median width obtained with all other 
$\gamma$ values. The most pronounced difference can be observed in Figure~\ref{fig:machine} at the $99\%$ confidence level, 
where the lower decile of the widths obtained with $\gamma = \infty$ surpasses the upper deciles of the widths obtained with 
$\gamma$ values of $2$, $3$ and $4$.

Conversely, when $\gamma$ is set to $1$ the median interval width is typically the smallest. However, this setting appears to 
result in an overly sensitive nonconformity measure, leading to some exceptionally wide PIs and, consequently, 
much larger mean widths, which are in many cases even larger than those produced with $\gamma = \infty$.

Increasing $\gamma$ reduces the sensitivity of the measure, leading to a decrease in the variability of PI widths, which is 
typically much higher above the median, and an increase in the median itself. Initially, the impact of the sensitivity reduction 
to the variability of the widths is more significant than the increase to their median value, resulting in, on average, tighter PIs. 
As $\gamma$ continues increasing, the effect of the variability reduction becomes less significant compared to the increase in the 
median, causing the mean PI widths to start expanding. Overall, the choice of $\gamma$ is dependent on the specific data set, 
with values in the range of $2$ to $4$ often resulting in the optimal balance for PI widths.

\section{Conclusion}\label{sec:conc}

This work developed a transductive Conformal Predictor based on the popular Gaussian Process Regression 
technique and defined a normalized nonconformity measure based on the variance produced by GPR.
The resulting approach retains all the desirable properties of GPR and guarantees well-calibrated PIs 
without assuming anything more than exchangeability of the data, as opposed to the much stronger 
assumptions of conventional GPR.

The presented experimental results highlight the superiority of the PIs generated by the proposed 
GPR-CP in comparison to existing approaches. In terms of calibration, the PIs of the original GPR 
and three of the four recalibration methods considered often exhibit much higher miscoverage rates than the 
corresponding significance level, rendering them potentially misleading. In contrast, 
GPR-CP consistently produces well-calibrated PIs regardless of the particular model and data characteristics. 
Conserning PI tightness, GPR-CP outperforms all other methods, with a substantial difference in mean widths 
on two of the four data sets, demonstrating the significant advantage of the proposed normalized nonconformity measure.

Immediate future directions of this work include the development of a GPR-CP variant based on sparse GP approaches 
(see e.g.~\cite{cao:sparcegpr}), 
that will enable the computationally efficient processing of large data sets. Additionally, further research will explore 
the development of new normalized nonconformity measures for regression CP, aiming at producing even tighter PIs.


%



\section*{Acknowledgment}
The author is grateful to Professors V.~Vovk and A.~Gammerman for useful discussions.

\begin{IEEEbiography}[{\includegraphics[width=1in,height=1.25in,clip,keepaspectratio]{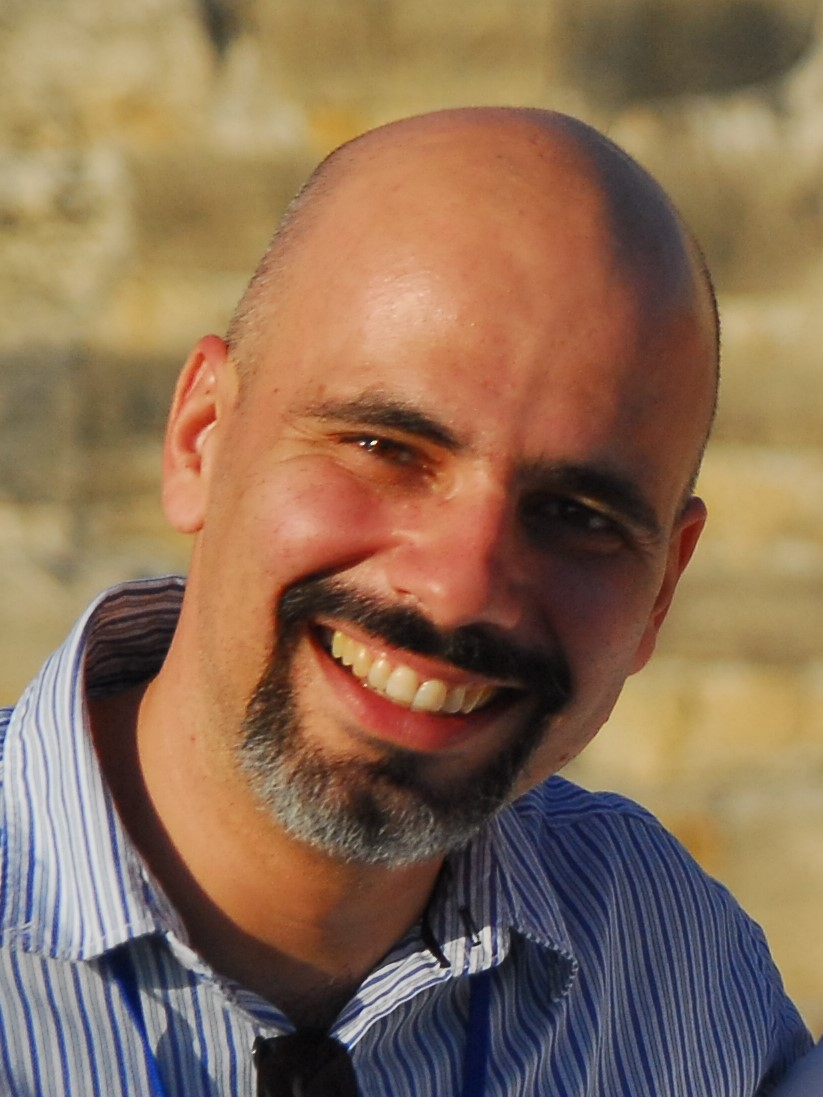}}]{Harris Papadopoulos}
is an Associate Professor at the Department of Electrical Engineering, Computer Engineering and Informatics of Frederick University, where he is heading the Computational Intelligence (COIN) Research Laboratory. He is also a Visiting Fellow of the Computer Learning Research Centre of Royal Holloway, University of London, UK. His research primarily focuses on the development of advanced Machine Learning techniques that provide probabilistically valid uncertainty quantification for each prediction under minimal assumptions, and on their application to a diverse array of challenges in various scientific fields. He has published more than 70 research papers in international refereed journals and conferences and co-edited the book “Measures of Complexity: Festschrift in Honor of Alexey Chervonenkis”, Springer, 2015. He has been involved in several funded research projects with a total budget of more than 6M euro, frequently taking the lead as the coordinator and initiator. Direct funding to his Lab over the past five years exceeds 1.5M euro. He served as an Associate Editor of the Evolving Systems Journal (Springer) from 2015 to 2018 and as guest editor of seven journal special issues. He also served as one of the main organizers of several international conferences and workshops.
\end{IEEEbiography}

\end{document}